 \documentclass[pmlr,twocolumn,10pt]{jmlr} 

\usepackage{booktabs}
\usepackage{siunitx}

\usepackage[switch]{lineno}

\theorembodyfont{\upshape}
\theoremheaderfont{\scshape}
\theorempostheader{:}
\theoremsep{\newline}

\jmlrvolume{LEAVE UNSET}
\jmlryear{2024}
\jmlrsubmitted{LEAVE UNSET}
\jmlrpublished{LEAVE UNSET}
\jmlrworkshop{Machine Learning for Health (ML4H) 2024} 

\title[Patient-Specific Models of Treatment Effects Explain Heterogeneity in Tuberculosis]{Patient-Specific Models of Treatment Effects Explain Heterogeneity in Tuberculosis}

 \author{
 \Name{Ethan Wu}\Email{etw46@pitt.edu}\\
 \addr Medical Scientist Training Program, Pitt School of Medicine
 \AND
 \Name{Caleb Ellington} \Email{cellingt@cs.cmu.edu}\\
 \addr Carnegie Mellon University
 \AND
 \Name{Ben Lengerich} \Email{lengerich@wisc.edu}\\
 \addr University of Wisconsin, Madison
 \AND
 \Name{Eric P. Xing} \Email{eric.xing@mbzuai.ac.ae}\\
\addr MBZUAI, CMU, Petuum Inc.
 }

\begin{document}

\maketitle

\begin{abstract}
Tuberculosis (TB) is a major global health challenge, and is compounded by co-morbidities such as HIV, diabetes, and anemia, which complicate treatment outcomes and contribute to heterogeneous patient responses. Traditional models of TB often overlook this heterogeneity by focusing on broad, pre-defined patient groups, thereby missing the nuanced effects of individual patient contexts. We propose moving beyond coarse subgroup analyses by using contextualized modeling, a multi-task learning approach that encodes patient context into personalized models of treatment effects, revealing patient-specific treatment benefits. Applied to the TB Portals dataset with multi-modal measurements for over 3,000 TB patients, our model reveals structured interactions between co-morbidities, treatments, and patient outcomes, identifying anemia, age of onset, and HIV as influential for treatment efficacy. By enhancing predictive accuracy in heterogeneous populations and providing patient-specific insights, contextualized models promise to enable new approaches to personalized treatment.
\end{abstract}
\begin{keywords}
tuberculosis, heterogeneity, context, personalized models
\end{keywords}

\paragraph*{Data and Code Availability}
Code available at \url{https://github.com/ethanwu2011/ContextualizedTB} and \url{https://contextualized.ml/}. Data was from the NIH TB Portals\footnote{\url{https://tbportals.niaid.nih.gov}}. 

\vspace{-2pt}

\paragraph*{Institutional Review Board (IRB)}
Our research does not require IRB approval.

\section{Introduction}
\label{sec:intro}
Tuberculosis (TB) remains a significant global health challenge, consistently ranking among the top ten causes of death worldwide \citep{chakaya_global_2021}. Managing TB in clinical practice is difficult due to the disease’s heterogeneity, with patients presenting varying degrees of drug resistance, co-morbidities, and socioeconomic factors that influence treatment outcomes. Despite advances in understanding TB pathophysiology, current antibiotic regimens remain standardized, failing to account for individual patient variability. This one-size-fits-all approach overlooks the interplay of patient-specific factors that impact treatment success \citep{fox_examining_2023, mdluli_tuberculosis_2015, zumla_tuberculosis_2015, dartois_anti-tuberculosis_2022}.

Co-morbidities such as HIV, Hepatitis B, Hepatitis C, Diabetes, and Anemia further complicate TB treatment. Anemia is prevalent among TB patients and has been associated with worse treatment outcomes \citep{isanaka_iron_2012, nagu_anaemia_2014}. For instance, immune suppression from HIV accelerates TB progression, while drug interactions between TB and HIV therapies demand careful management \citep{zumla_tuberculosis_2015, dartois_anti-tuberculosis_2022}. HIV patients are 18 times more likely to develop active TB, with TB responsible for one-third of HIV-related deaths globally \citep{haiminen_impact_2022}. These challenges highlight the need for personalized treatment strategies and a shift away from one-size-fits-all approaches. 

Recently, multi-modal TB patient data has provided new opportunities and challenges for modeling this disease. Here, we use the TB Portals dataset, which includes treatments and outcomes along with clinical, demographic, and imaging contexts, making it well-suited to identify sources of heterogeneity in TB. However, methods for accommodating heterogeneous multi-modal data in statistically principled models are still nascent. A number of studies have previously interrogated the TB Portals dataset for drivers of heterogeneity in TB, but apply off-the-shelf machine learning methods that make explicit assumptions about data homogeneity, making them ill-suited for highly heterogeneous data \citep{fox_examining_2023, barros_benchmarking_2021, rosenthal_tb_portals_2017}. 

\subsection*{Approach}

\noindent
We define the problem of heterogeneity and multi-modality in treatment effect models through the lens of personalized modeling, aiming to provide a solution to problems in both TB and other heterogeneous and multi-faceted diseases. Personalized models represent heterogeneity by using sample-specific distributions: For each patient indexed by $i$, their outcome $Y_i$ and treatment $X_i$ are drawn from a distribution specific to that patient $P_i(Y \mid X)$. An enabling assumption is that all $P_i$ belong to the same parametric family, i.e., $Y_i \sim P(Y_i \mid X_i, \theta_i)$. Now heterogeneity among patients can be quantified and explained by estimating sample-specific parameters $\theta_i$. While some methods try to infer each sample-specific parameter independently, this leads to high variance estimators with poor generalization, and most methods make use of side information $C$ (i.e. metadata or context) as a proxy for the differences among sample-specific distributions \citep{al-shedivat_contextual_2020, hastie_varying-coefficient_1993, wang_bayesian_2022,  ellington_contextualized_2024, deuschel_contextualized_2023, lengerich_contextualized_2023, lengerich_automated_2022, lengerich_learning_2019, lengerich_personalized_2018, lengerich_discriminative_2022}.
Under this view, 
$$P(Y \mid X, C) = \int_\theta d\theta P(Y \mid X, \theta)P(\theta \mid C),$$
where $P(Y \mid X, \theta)$ describes a sample-specific model, and $P(\theta \mid C)$ relates context to model parameters. 
\begin{figure}[tbp]
    \floatconts
    {fig:context_diagram}
    {\caption{Contextualized Modeling Diagram}}
    {\includegraphics[trim=0 20 0 10, clip, width=1.0\linewidth]{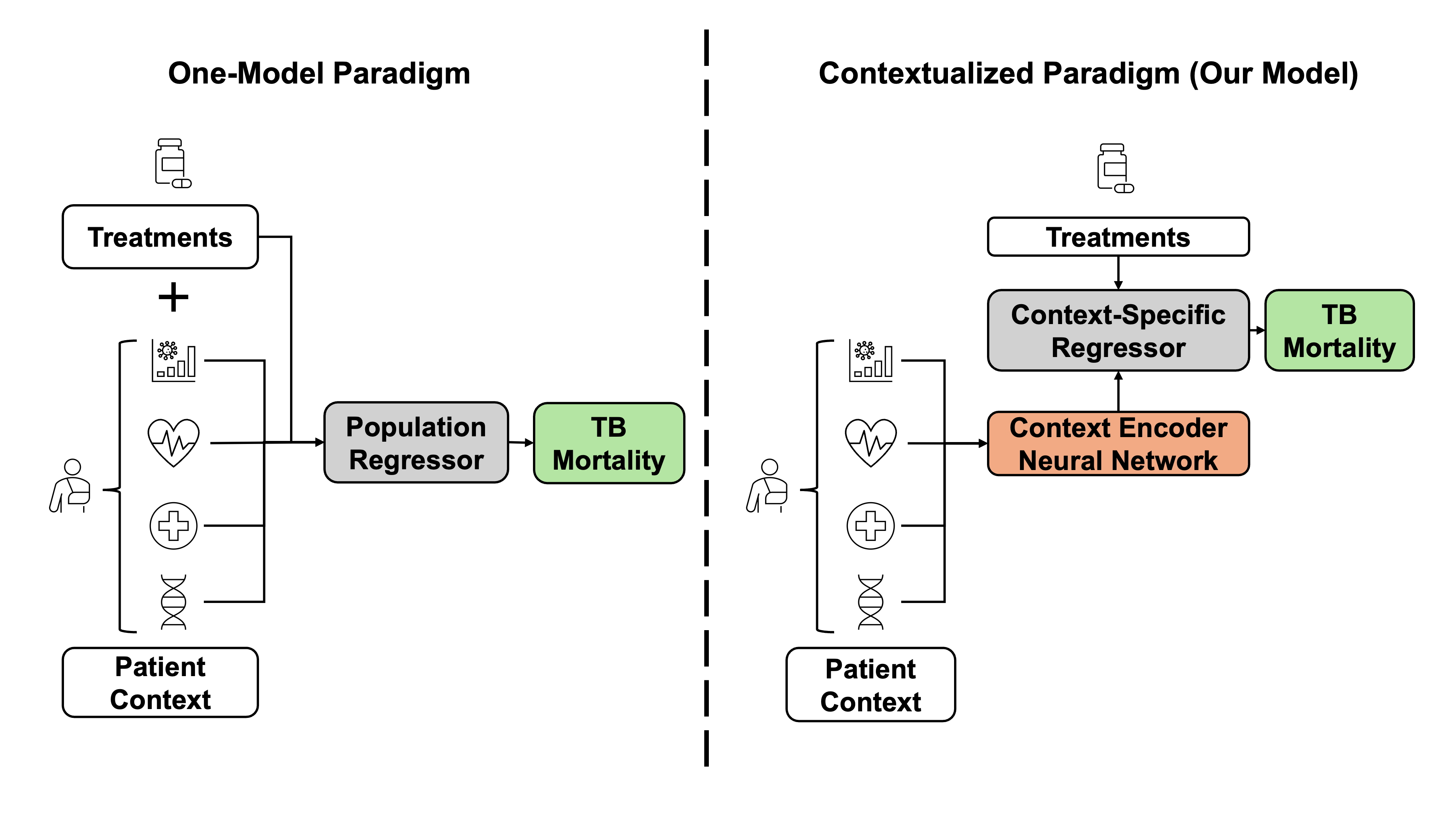}}
\end{figure}
From this perspective, multi-modal data and heterogeneity are not orthogonal issues, but are synergistic.  
Recently, contextualized machine learning \citep{al-shedivat_contextual_2020, lengerich_contextualized_2023, ellington_contextualized_2024} was proposed to exploit this synergy by using deep learning to encode context $P(\theta \mid C)$, 
removing assumptions on the functional form of the context encoder.
In practice, these approaches also assume the context encoder $P(\theta | C)$ is deterministic, and hence  
$P(Y | X, C) = P(Y | X, \theta = f(C))$.

In this study, we use contextualized machine learning to analyze heterogeneous treatment effects in the TB Portals dataset by inferring patient-specific treatment models. For context $C$, treatments $X$, and mortality $Y$, contextualized logistic regression decomposes treatment effect models into three components
\begin{equation}
    \text{logodds}(Y | X, C) = X\beta + X\beta(C) + \mu(C) 
    \label{eq:contextual_effects}
\end{equation}
describing the context-invariant or \textit{population} treatment effects $\beta$ (homogeneous treatment effects), the context-dependent treatment effects $\beta(C)$ (heterogeneous treatment effects), and the direct effects of context $\mu(C)$ (homogeneous context effects) (Fig. \ref{fig:context_diagram}). 
This decomposition allows us to understand which contexts directly affect patient outcomes, which have indirect effects by modifying treatment dose-response, and which treatments context-invariant effects.
While previous works have explored heterogeneity and homogeneity in data by visualizing the inferred sample-specific model parameters \citep{deuschel_contextualized_2023, lengerich_automated_2022}, we develop principled tests to quantify the significance and degree of each of these effects.
Specifically, we perform many bootstrap estimates of the contextualized model parameters on the centered data, which allows us to test the consistency of the estimator using a one-sided t-test on the sign of the estimates for each parameter.
For the homogeneous effect estimates $\widehat{\mu}(C)$ and $\widehat{\beta}$, this is sufficient.
For the heterogeneous effects $\widehat{\beta}(C)$, we assume the true $\beta(C)$ is monotonic over $C$ and instead test the difference between the high and low end of the observed range of $C$.

\section{Results}

We investigate TB heterogeneity by inferring contextualized models of patient-specific treatment effects \citep{lengerich_contextualized_2023, ellington_contextualized_2024}, achieving accuracy comparable to or better than traditional statistical models and black-box models like regression forests and neural networks (Table \ref{tab:model_accuracy_table}). We systematically identify and quantify heterogeneity in treatment effects related to patient contexts, such as co-morbidities (Figures \ref{fig:heterogeneity heatmap}, \ref{fig:Context&Treatment}). Importantly, we distinguish between heterogeneous treatment dose-responses and direct context effects on outcomes (Figure \ref{fig:context_effects}), 
avoiding the need for extensive pairwise tests. 

\subsection{Contextualization Improves Predictive Accuracy of Treatment Outcomes}

\noindent
To assess the necessity of context in modeling patient outcomes and the accuracy of our model, we benchmarked models for predicting survival/mortality outcomes in three regimes: (1) using only treatment information, (2) using both treatment and additional contextual data (clinical and imaging) in a concatenated feature vector, and (3) using our contextualized model that predicts changes in treatment efficacy based on context (Table \ref{tab:model_accuracy_table}). Accounting for context improves accuracy in all cases, with contextualized models achieving 83.7\% predictive accuracy, particularly improving the prediction of mortality. While accuracy and interpretability are often considered mutually exclusive, here contextualized models improve accuracy and provide mechanistic explanations for differences in treatment effects.
\begin{table}[bp]
\floatconts
  {tab:model_accuracy_table}
  {\caption{Accuracy of overall survival prediction, and class-specific accuracy for mortality.}}
  {\renewcommand{\arraystretch}{0.7} 
  \begin{tabular}{lll}
  \toprule
  \bfseries Model Type & \textbf{Overall} & \textbf{Mortality} \\
                              & \textbf{Accuracy} & \textbf{Accuracy} \\
  \midrule
  \bfseries Without Context & \\
  \quad Logistic Regression & 0.803 & 0.193 \\
  \quad GBDT & 0.807 & 0.199 \\
  \midrule
  \bfseries With Context & \\
  \quad Logistic Regression & 0.828 & 0.344 \\
  \quad GBDT & 0.834 & 0.358 \\
  \midrule
  \bfseries Context + Encoder & 0.837 & 0.386 \\
  \bottomrule
  \end{tabular}}
\end{table}

\subsection{Contextualization Reveals Direct Context and Treatment Effects}
Exploring the direct effects of context $\mu(C)$ (Fig. \ref{fig:context_effects}), we find that co-morbidities such as HIV and Anemia, as well as onset age and employment status, have both homogeneous and heterogeneous effects, showing significant direct effects on survival likelihood (Fig. \ref{fig:context_effects}), while also significantly influencing the heterogeneity of treatment effects (Fig. \ref{fig:heterogeneity heatmap}). At the same time, socioeconomic and social factors such an unemployment, disabled, and student identity also show decent direct effects on survival likelihood. 

\begin{figure}[tbp]
    \floatconts
    {fig:context_effects}
    {\caption{Homogeneous Effects of Context}}
    {\includegraphics[trim=0 20 0 10, clip,width=1.0\linewidth]{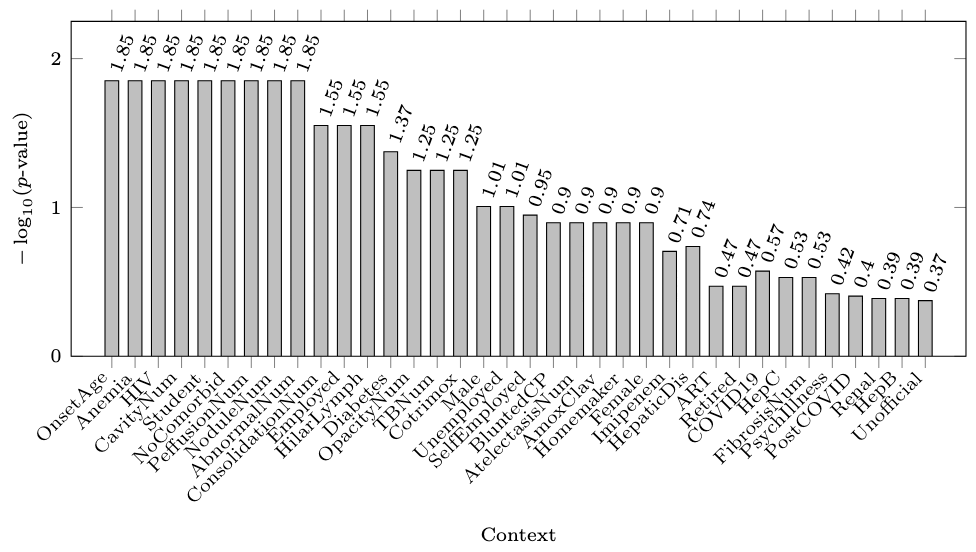}}
\end{figure}
\begin{figure}[tbp]
    \floatconts
    {fig:heterogeneity heatmap}
    {\caption{Heterogeneous Treatment Effects. The significance of the change in treatment efficacy over context, visualized as $-\text{log}(\text{p-val})$}}
    {\includegraphics[trim=0 20 0 10, clip,width=1.0\linewidth]{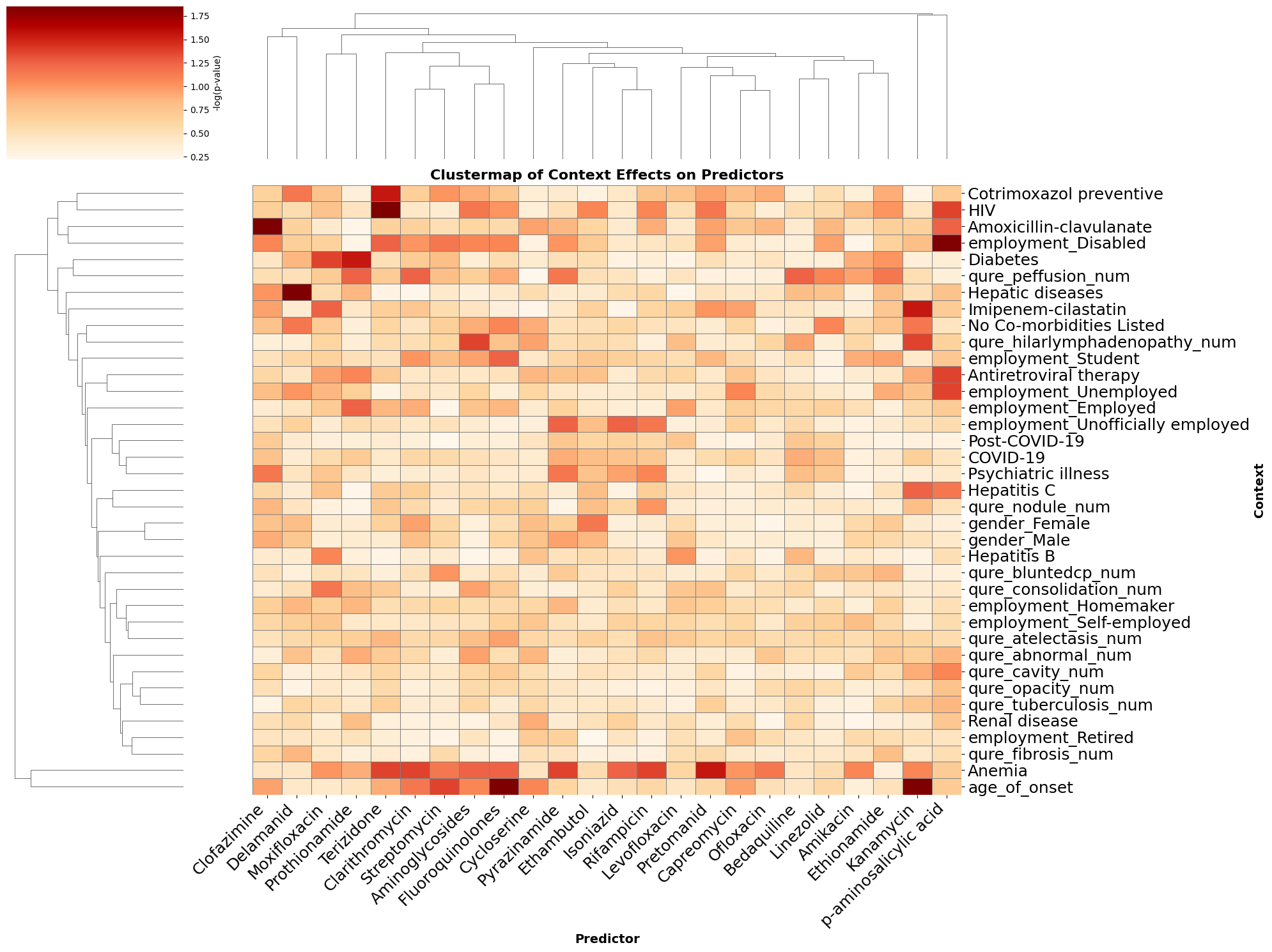}}
\end{figure}
\subsection{Contextualized Models Reveal Critical Context-Treatment Interactions}
\noindent
Our analysis of treatment heterogeneity ($\beta(C)$) across patient contexts uncovers key interactions that have clinical implications. Anemia significantly alters the efficacy of treatments like Aminoglycosides, Clarithromycin, and Pyrazinamide. Additionally, the analysis also identified age of onset as a significant predictor for several drugs, including Kanamycin and Fluoroquinolones. This suggests that younger or older patients may respond differently to these treatments, which has important implications for optimizing drug regimens across age groups. (Fig. \ref{fig:heterogeneity heatmap}).

\begin{figure}[tbp]
  \label{fig:Context&Treatment}
    \subfigure[Treatments most affected by patient context][c]{\label{fig:circle2}%
      \includegraphics[trim=0 20 0 10, clip,width=1.0\linewidth]{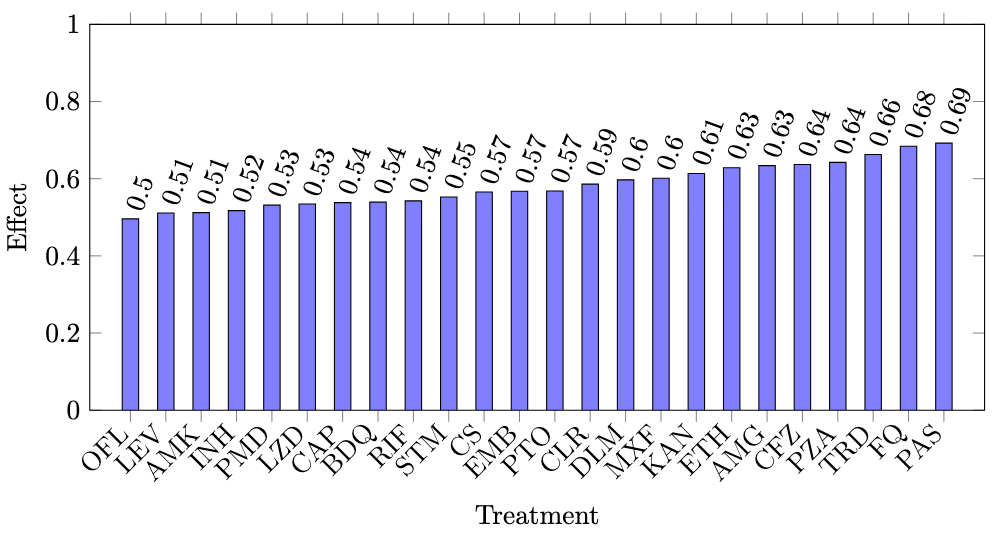}}%
    \qquad
    \subfigure[Contexts with the broadest effects on treatments][c]{\label{fig:square2}%
      \includegraphics[trim=0 20 0 10, clip,width=1.0\linewidth]{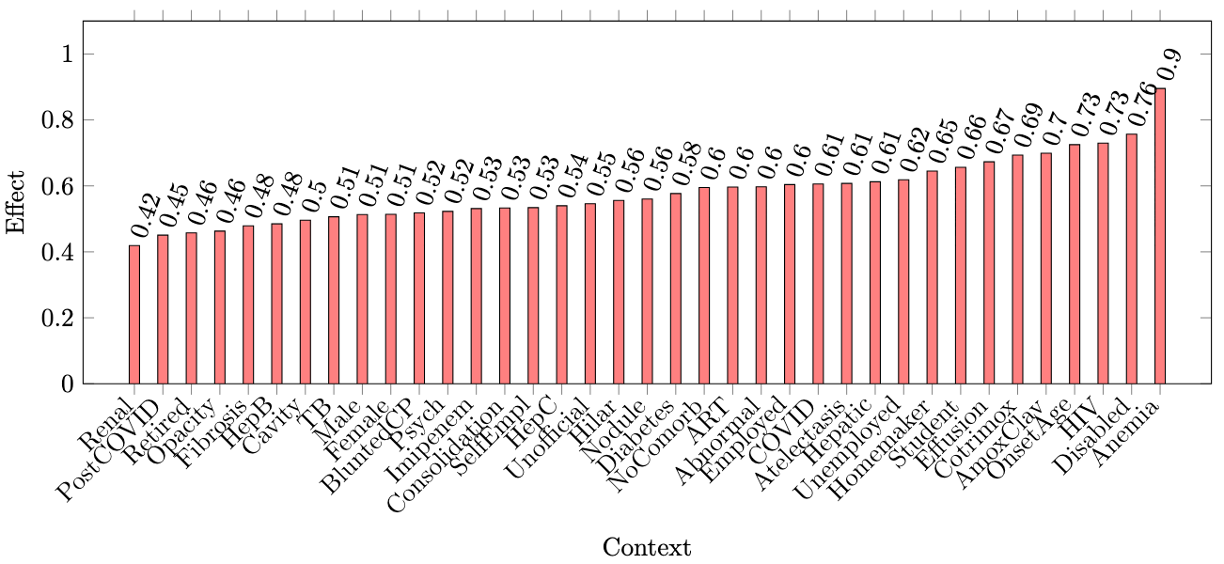}}
  \caption{Heterogeneous Effects of Treatments}
\end{figure}

Benchmarking treatment sensitivity across patient contexts provides insights for clinical decision-making. Terizidone (TRD), Fluoroquinolones (FQ), and para-aminosalicylic acid (PAS) showed increased sensitivity to patient variability, highlighting the need for personalized prescribing. The analysis (Fig. \ref{fig:Context&Treatment}) also showed that Anemia, often overlooked in TB treatment, emerged as a significant context, suggesting it deserves more attention in clinical trials. 
\section{Discussion}
\label{sec:math}
In this study, we introduced a contextualized machine learning approach to address the challenge of predicting tuberculosis (TB) treatment outcomes in the presence of diverse patient-specific contexts. Current statistical methods often fall short when it comes to identifying biomarkers and treatment strategies that are effective across clinically diverse patient populations. Our contextualized model offers a solution by learning a cohesive, sample-specific representation of latent patient states from patient context, capturing the interactions between patient-specific factors, treatments, and treatment outcomes through estimated patient-specific treatment effects. Moreover, contextualized models excel in estimating context-specific effects when dealing with observational data that is often heterogeneous. By sharing information across samples and allowing for sample-specific variation, our approach navigates the trade-off between model generalization and specificity. This is important for TB, where controlling for all conditions and contexts would otherwise lead to data scarcity and unreliable inferences.

Our results show that contextualization enhances predictive accuracy and identifies subpopulations with distinct patient-specific factors to highlight specific treatment plans (Figure \ref{fig:heterogeneity heatmap}) that could serve as personalized recommendations. 

Our findings have direct clinical implications. Anemia emerged as a key factor influencing the effectiveness of TB treatments, including Streptomycin and Pyrazinamide, suggesting that this often-overlooked co-morbidity should be closely monitored to ensure treatment efficacy. Similarly, HIV status significantly affected Terizidone outcomes, emphasizing the need for careful management of HIV-TB co-infected patients. The study also highlights the age of onset as a critical factor, with treatments like Kanamycin, Streptomycin, and Fluoroquinolones showing age-specific variability, particularly in elderly patients who may require tailored regimens due to differences in drug metabolism and immune function.

Contextualized models provide a more nuanced understanding of TB treatment outcomes in heterogeneous patient populations.
The ability to integrate multi-modal data into the treatment modeling process represents a significant advancement in the field of personalized medicine, 
enabling treatment plans which are tailored to the unique characteristics of each patient,
while also capturing more views on patient health by providing new opportunities for data collection and utilization beyond traditional modeling approaches,
both of which ultimately enhance the likelihood of successful patient outcomes.

\newpage

\bibliography{main}

\end{document}